\documentclass[journal]{IEEEtran}
\usepackage{algorithm}
\usepackage{algpseudocode}
\usepackage{amsmath}
\usepackage{graphicx}
\usepackage{multirow}
\begin{document}
\title{Using dynamic routing to extract intermediate features for developing scalable capsule networks}

\author{
    \IEEEauthorblockN{Bodhisatwa Mandal\IEEEauthorrefmark{1}, Swarnendu Ghosh\IEEEauthorrefmark{1}, Ritesh Sarkhel\IEEEauthorrefmark{2}, Nibaran Das\IEEEauthorrefmark{1}, Mita Nasipuri\IEEEauthorrefmark{1}}
   \\
    
    \IEEEauthorblockA{\IEEEauthorrefmark{1}Jadavpur University, Kolkata, 700032, WB,India,
    \\ bodhisatwam@gmail.com, \{swarnendughosh.cse.rs, nibaran.das, mita.nasipuri\}@jadavpuruniversity.in}
    \\ \IEEEauthorblockA{\IEEEauthorrefmark{2}Ohio State University, Columbus, OH 43210, USA}
    \\ sarkhelritesh@gmail.com}

\maketitle

\begin{abstract}
Capsule networks have gained a lot of popularity in short time due to its unique approach to model equivariant class specific properties as capsules from images. However the dynamic routing algorithm comes with a steep computational complexity. In the proposed approach we aim to create scalable versions of the capsule networks that are much faster and provide better accuracy in problems with higher number of classes. By using dynamic routing to extract intermediate features instead of generating output class specific capsules, a large increase in the computational speed has been observed. Moreover, by extracting equivariant feature capsules instead of class specific capsules, the generalization capability of the network has also increased as a result of which there is a boost in accuracy.
\end{abstract}


\IEEEpeerreviewmaketitle

\section{Introduction}
Convolutional Neural Networks(CNNs) have been around for almost two decades now. Since introduction in 1998 for digit classification problem \cite{lecun1998gradient} it took almost 14 years to improve them for more complicated problems \cite{krizhevsky2012imagenet,roy2017handwritten}. CNNs have grown deeper and got better in the field of visual recognition \cite{szegedy2015going,he2016deep,huang2017densely} . However it suffered from a key issue of generating invariant features. It mainly focused on the amount of activations with respect to different features without focusing on the correlation among the presence of various features. A notable attempt has been made to address this issue in the capsule networks. In capsule networks, intermediate activations are represented in the form of vectors or primary capsules \cite{sabour2017dynamic}. Moreover, unlike the normal CNNs, the activations in the output layer are also represented in the form of vectors or output capsules(as defined in \cite{sabour2017dynamic}). Capsule networks also implement a technique called ``dynamic routing'' that measures the agreement among the primary capsules and generate the output capsules. Dynamic routing involves a weighted summation of primary capsules where the weights are iteratively updated during the forward proportional to the similarity between the individual activations and the combined activation. In the native work the network was demonstrated to perform well for 10 class problems like MNIST \footnote{http://yann.lecun.com/exdb/mnist/}, CIFAR-10 \footnote{https://www.cs.toronto.edu/~kriz/cifar.html}, SVHN \footnote{http://ufldl.stanford.edu/housenumbers/} and so on. 
\begin{figure}[b]
   \centering
     \includegraphics[width = 0.48\textwidth]{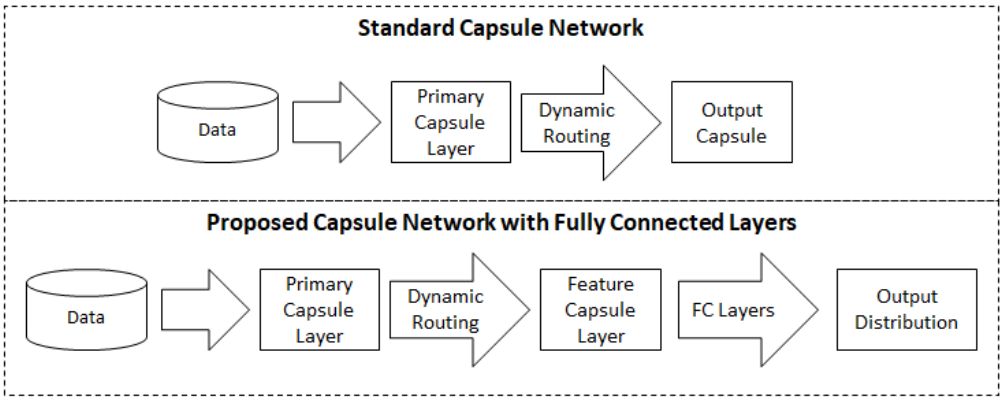}
    \caption{Summary of the proposed system}
    \label{fig:system}
\end{figure}
Capsule networks have also been shown to improve upon more complex networks like AlexNet for Indic character recognition \cite{mandal2018capsule}. Another approach proposed by the same authors of capsule network demonstrated the use of matrix capsules \cite{hinton2018matrix} instead of vector capsules to improve upon the performance. However, while implementing on problems with higher number of classes it was observed that capsule networks take a lot of time to train. Not only that, it was also seen the performance severely degrades as the number of classes increases. The degradation of performance of capsule networks for more complicated dataset was shown in the works of \cite{xi2017capsule}. Thus, capsule networks by itself is not scalable for more complicated problems. In its native form the dynamic routing algorithm depends directly on the number of classes. While for low number of classes it is easier to model agreement between primary capsules, for complicated problems, the number of interrelationships increases considerably. Thus normal capsule networks tend to over-fit the data. In the current work we propose to use capsule networks for feature extraction rather than generating class specific capsules. By using capsules for generating intermediate features we abolish the dependency of dynamic routing on the number of classes. Moreover the number of features generated by dynamic routing can be much lesser than the number of classes thus preventing the network from over-fitting while learning more general features in the process.  A brief overview of the proposed system is shown in fig. \ref{fig:system}. The proposed approach is tested on three Indic digit database \cite{obaidullah2018phdindic_11} namely ``Bangla'', ``Devanagari'', and ``Telugu'', as well as more complicated datasets with a higher number of class such as ``Bangla Basic'', ``Bangla Compound''.
\begin{figure*}[t]
    \centering
     \includegraphics[width = 0.7\textwidth]{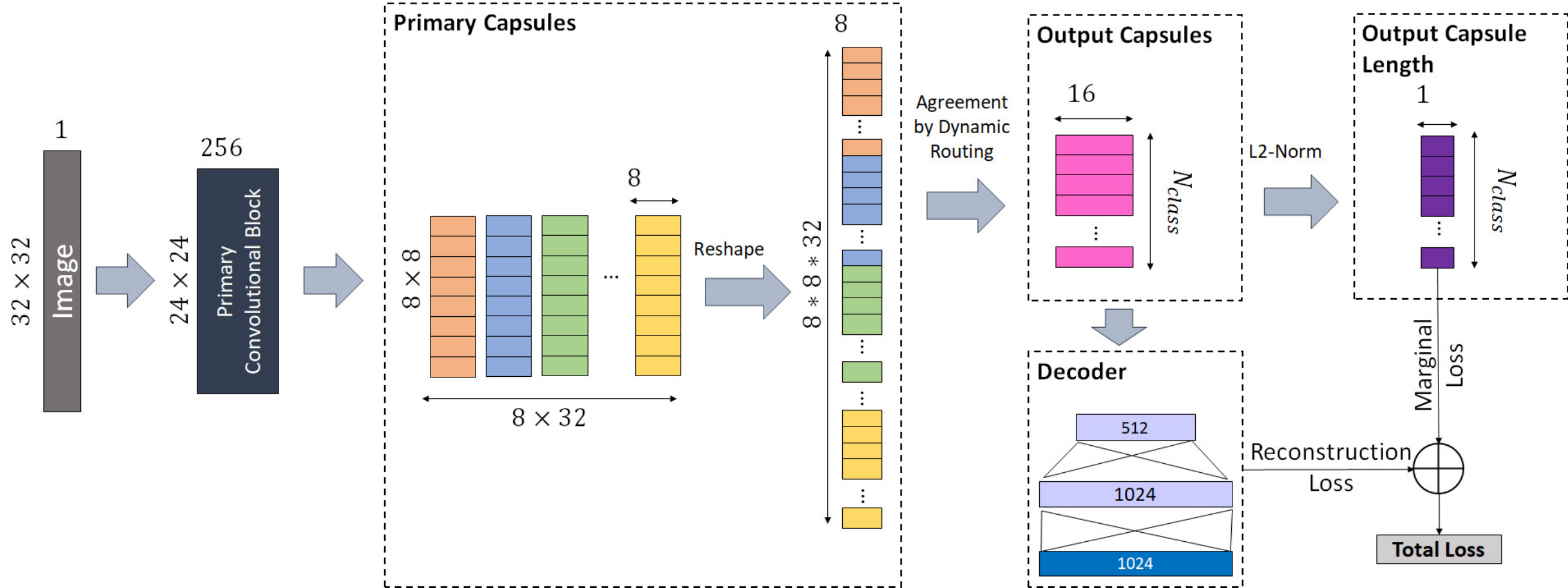}
    \caption{Original capsule network}
    \label{fig:caps}
\end{figure*}
\section{The capsule network}
While traditional CNNs are invariant to object positions and orientations, capsule networks aim to bring equivariance of pose vectors. In capsule networks, the successive layers get a higher activation when the kernels in the previous layers agree to the same decision. A schematic diagram of capsule network is shown in fig. \ref{fig:caps}. It mainly consists two different layers, the primary capsule layer and the output capsule layer. The primary capsule layer groups together outputs from multiple convolutions into a single capsule unit. The output of primary layer is fed into output capsule layer via dynamic routing of capsules. Each of the vectors of the output capsule layer corresponds to a single output class. Length of the vectors corresponds to the likelihood of its corresponding class.

\subsection{Primary Capsule Layer}
Capsule networks start with convolution layers that convert image samples to group of activation tensors. These tensors are converted to primary capsules in the primary capsule layer. Considering single channel inputs, $256$ number of  $9\times 9$ kernel is convoluted with a stride of $1$ to obtain a $256$-dimensional feature for the image. Unlike the scalar activations of a traditional convolutional layer, primary capsule are in the form of a $8$-dimensional vectors generated by $8$ number of $9\times 9$ kernels convoluted with a stride of $2$. By have $32$ such groups of kernels, $32$ blocks of primary capsules are generated. These blocks are reshaped in a way such that all the 8-dimensional primary capsules are lined up to form a $8 \times N_{PC}$ tensor where $N_{PC}$ is the total number of primary capsules.
 
\subsection{Output Capsule Layer}
The output capsule layer takes the primary capsules from the previous layer and using dynamic routing, produces output capsules. Each capsule of the output capsule layer is a vector describing a single class. Usually, the output layer of a fully connected neural network produces $N_{class} \times 1$ as the output, $N_{class}$ being the number of classes for a given dataset. However, with respect to capsule networks, the output capsule layer is of the dimension $N_{class} \times 16$ where each class is represented by a $16$ dimensional output capsule. Instead of providing with a scalar representation of the likelihood of each class, the information is encapsulated in a vector of dimension $16$. Length of the output vector denotes the likelihood of the presence of the corresponding class.

\begin{figure*}[t]
    \centering
    \includegraphics[width = 0.7\textwidth]{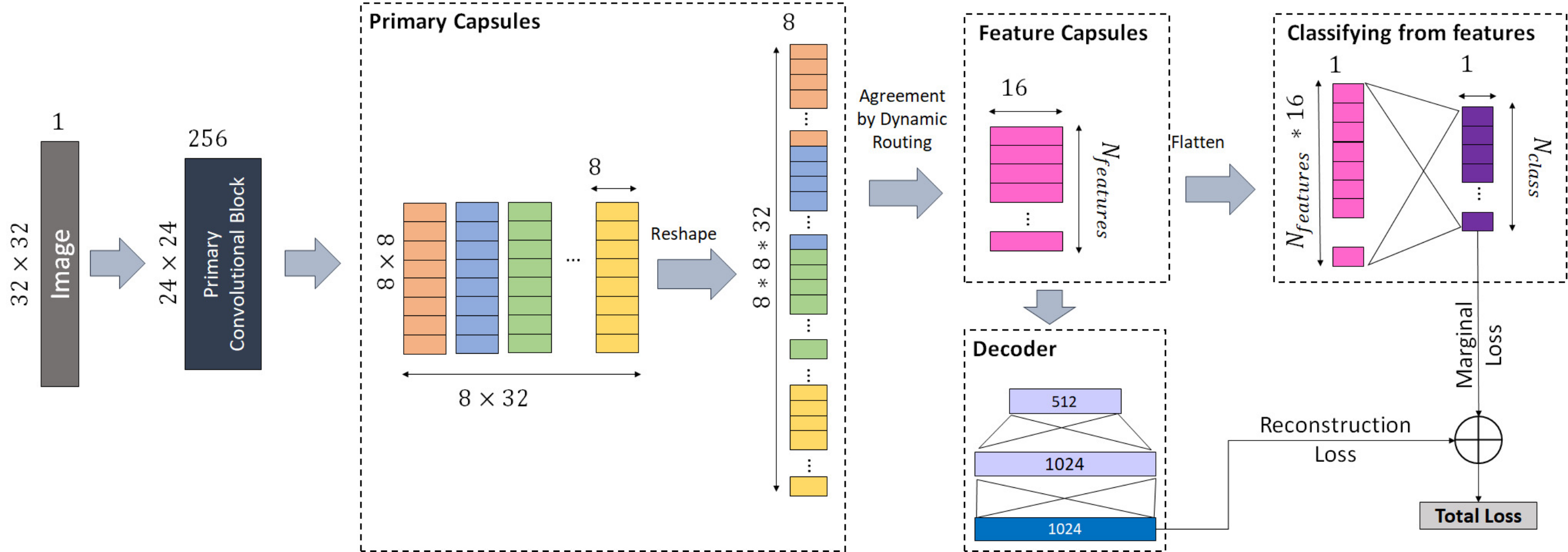}
    \caption{The proposed network}
    \label{fig:proposed}
\end{figure*}

\subsection{Dynamic Routing} 
The output capsules are obtained from the primary capsules by the operation of dynamic routing. The trainable weights of dynamic routing, that is, $W$ are used to determine the individual opinions of every capsule. Considering $ i \in [1,N_{PC}]$ to be the index of the 8-dimensional primary capsules of dimension and $j \in [1,N_{class}]$ to be the index of the 16 dimensional output capsules, $W_{ij}$ has the dimension $ 8 \times 16$. The individual opinion of the primary capsule $i$ regarding the output capsule $j$ is given by : 
\begin{equation}
\hat{u}_{j|i} = u_i \times W_{ij}^{DC}    
\end{equation}
where the $i$-th primary capsule is denoted by $u_i$ . For every primary capsule $i$, we get a output block of shape $N_{class} \times 16$.
For the operation of dynamic routing, another type of weight is considered of dimension $N_{PC} \times N_{class}$, called routing weights, $b$. They are used in combining individual opinions to form the final output capsules. However, unlike $W$, these weights are learned not by backpropagation, but are learned through dynamic routing during forward pass, depending on the degree of agreement of the individual opinions with the combined output. These weights are initialized as zeros on the start of every forward pass. The coupling coefficients $c_{ij}$ is given by :
\begin{equation}
c_{ij} = \frac{exp(b_{ij})}{\sum_k exp(b_{ik})}
\end{equation}
These coupling coefficients are used for combining the individual opinions $\hat{u}_{j|i}$ to form the combined output capsule. The $j$-th combined output capsule is obtained by squashing $s_j$ is given by:
\begin{equation}
  s_j = \sum_i c_{ij}\hat{u}_{j|i}  
\end{equation}
A non-linear "squashing" function is used to ensure that shorter vectors get shrunk to nearly zero length and longer vectors get shrunk to a length slightly below 1. The output capsule $v_j$ is given by :
\begin{equation}
   v_j = \frac{||s_j||^2}{1+||s_j||^2} \frac{s_j}{||s_j||} 
\end{equation}
A simple dot product calculates the agreement between individual output capsules $\hat{u}_{j|i}$ and the squashed combined output capsules $v_j$. The individual capsules having more agreement with the combined output are given higher preference. This is done by updating the  $b_{ij}$ as : 
\begin{equation}
   b_{ij} = b_{ij} + \hat{u}_{j|i}.v_j 
\end{equation}

\subsection{Loss Function}
 The loss function is divided into two parts, the margin loss for object existence and mean square loss with respect to the generated images from the output capsules.The marginal loss for object $k$ is given by :
\begin{multline}
L_k = T_k\ max(0,m^{+} - ||v_k||)^2 + \\
	 \lambda\ (1-T_k)\ max(0,||v_k||-m^{-})^2
\end{multline}
 Here, $Tk = 1$ iff a object of class $k$ is present. The upper and lower bounds $m_+$ and $m^−$ are set to 0.9 and 0.1 respectively.$\lambda$ is set to 0.5.
 
 \subsection  {Decoder network for regularization} 
 The decoder network is a series of fully connected layers to reconstruct the original input. The output capsule layer is masked out to convert all capsules but the one corresponding to the label to zero. The masked out output capsules are sent to the decoder for reconstruction. The masking helps in creating class-specific reconstructions during the test time. The reconstruction loss is minimized along with the margin loss so that the model does not overfit on the training dataset. To prevent the reconstruction loss from dominating the margin loss, the former is scaled down using a factor of 0.0005.

\section{The Proposed Approach}
The main drawback of capsule network is the computation time for doing the dynamic routing. The number routing weights $b$ is directly proportional to the number of classes. Traditionally, capsule networks were shown to perform only with smaller number of classes. But classification of higher class datasets become extremely time consuming to the point that it is virtually impractical. It has also been observed that with an increasing number of classes, the performance of capsule networks drop significantly. With an increasing number of classes, analyzing agreement among capsules become more complicated. The network tends to overfit the data and hence resulting in an overall degradation in performance. 

\subsection{Capsule Connections as Feature Extractor}
To overcome the above mentioned issues, a modification to the capsule network has been proposed. Instead of performing dynamic routing to obtain class-specific output capsules, it has been used to map the inputs to an intermediate feature space. Thus, the dynamic routing phase becomes invariant to the number of classes, rather it depends only on the number of features extracted. A schematic diagram of our proposed is shown in fig. \ref{fig:proposed}.
In our proposed model, the capsule network gives feature specific capsules instead of class specific capsules. Samples in a dataset may be represented using features fewer than the number of classes in that dataset. This enables us to target higher class datasets using smaller number of features. Since majority of the time of execution is taken up by computation of features using capsule network, keeping the number of features same, increase in the number of output classes does not have major impact on the time taken to train the network. Furthermore by limiting the number of features the generalization capability of the network is also boosted, thus resulting in an improved performance. In the original capsule networks equivariance among capsules were measured to analyze agreement among them regarding the presence of one of the output classes. In the present work, the agreement among capsules are measured with respect to presence of some specific features from which the output classes are inferred through another fully connected layer.

\subsection{Feature Capsule Layer}
Instead of a final output capsule layer, we use an intermediate feature capsule layer that will create feature capsules from original primary capsules. For input of size $N_{PC} \times 8$, feature capsule layer creates feature capsules of size $N_{features} \times 16$ where $N_{features}$ should ideally be less than $N_{class}$ for a gain in speed as well as reduction of memory footprint compared to normal capsule network. These feature capsules are flattened and passed into a fully connected output layer to compute the class specific likelihood. If the capsule features are represented by $v_f$ for $f\in [1,N_{features}]$ and the output probability for class $j$ for $j\in [1,N_{class}]$ is written as, 
\begin{equation}
    v'_j = \frac{\exp{FC_j(v_f)}}{\sum_k{\exp{FC_k(v_f)}}}
\end{equation}
where $FC_{j}$ is the $j-th$ output of a fully connected layer.
\subsection{Regularization using Feature Capsules}
 Unlike native capsule networks, the decoder tries to reconstruct the input from the feature capsules rather than the output capsules. These reconstructed images are used as a regularization method by minimizing the reconstruction loss, thereby preventing the network from over-fitting. Just like normal capsule network, the reconstruction loss is scaled down by a factor of 0.0005 so that the margin loss is not dominated.

\subsection{The modified loss function}
Similar to original capsule networks, the margin loss for a class is computed. For class $k$ the margin loss is given by,
\begin{multline}
M_k = T_k \max(0,m^+-v'_k)^2 + \\ \lambda (1-T_k)  \max(0,v'_k - m^-)^2. 
\end{multline}
Here, $T_k = 1$ iff a sample of a class $k$ is present, else $T_k =0$. The upper and lower bounds $m^+$ and $m^-$ are set to 0.9 and 0.1 respectively and $\lambda$ is set to 0.5 \newline
The decoder network tries to reconstruct the original samples from intermediate feature capsules. Unlike the capsule networks, the decoder recieves the entire set of feature capsules without any form of masking. The reconstruction loss is given by simple mean square error between the input and the reconstructed image. The reconstruction loss is given by,
\begin{equation}
    R = MSELoss(X,X')
\end{equation}
where, $X$ is the input image and $X'$ is the reconstructed image. Finally, the net loss is calculated by ,
\begin{equation}
    L_k = M_k + \beta R
\end{equation}
Here $\beta$ is the scale down factor that prevents the reconstruction loss from dominating over the margin loss. In our experiments, the value of $\beta$ is taken as 0.0005, same as the one used in original capsule networks.

\section{Experimental Setup}
\begin{table}[b]
\begin{center}
	\caption{Dataset Descriptions}
	\label{tab:Dataset}
	\resizebox{0.48\textwidth}{!}{%
	\begin{tabular}{l  c  c  c} 
		\hline
		\textbf{Dataset} & \textbf{Classes} & \textbf{Train Set} &\textbf{ Test Set}\\ [1.0ex] 
		\hline\hline
		\textbf{\textit{Bangla Numeral}} & 10 & 4000 & 2000\\ [0.5ex]
	    \textbf{\textit{Devangari Numeral}} & 10 & 2000 & 1000\\ [0.5ex]
	    \textbf{\textit{Telugu Numeral}} & 10 & 2000 & 1000\\ [0.5ex]
		\textbf{\textit{Bangla Basic Character}} & 50 & 12000 & 3000\\ [0.5ex]
	    \textbf{\textit{Bangla Compound Character}} & 199 & 33282 & 8254\\ [0.5ex]
		\hline
	\end{tabular}%
	}
\end{center}
\end{table}

\begin{table}[thb]
\centering
\caption{Training Time (in seconds) per Sample per Epoch for Numeral Datasets}
\label{tab:time-per-sample}
\resizebox{0.48\textwidth}{!}{%
\begin{tabular}{lcccccc}
\hline
\multirow{2}{*}{\textbf{Dataset}} & \multirow{2}{*}{\textbf{\begin{tabular}[c]{@{}c@{}}Capsule\\ Network\end{tabular}}} & \multicolumn{5}{c}{\textbf{$N_{features}$}} \\[0.5ex]
 &  & \textbf{2} & \textbf{4} & \textbf{6} & \textbf{8} & \textbf{10} \\\hline\hline 
\textit{\textbf{Bangla Numeral}} & 0.0677 & 0.0655 & 0.0657 & 0.0670 & 0.0672 & 0.0677 \\[0.5ex]
\textit{\textbf{Devanagari Numeral}} & 0.0677 & 0.0655 & 0.0657 & 0.0670 & 0.0672 & 0.0677 \\[0.5ex]
\textit{\textbf{Telugu Numeral}} & 0.0677 & 0.0655 & 0.0657 & 0.0670 & 0.0672 & 0.0677 \\[0.5ex]
\hline
\end{tabular}%
}
\end{table}

\begin{table}[thb]
\centering
\caption{Memory Consumption (in megabytes) per Sample for Numeral Datasets}
\label{tab:memory-consumption}
\resizebox{0.48\textwidth}{!}{%
\begin{tabular}{lcccccc}
\hline
\multirow{2}{*}{\textbf{Dataset}} & \multirow{2}{*}{\textbf{\begin{tabular}[c]{@{}c@{}}Capsule\\ Network\end{tabular}}} & \multicolumn{5}{c}{\textbf{$N_{features}$}} \\[0.5ex]
 &  & \textbf{2} & \textbf{4} & \textbf{6} & \textbf{8} & \textbf{10} \\\hline\hline 
\textit{\textbf{Bangla Numeral}} & 719 & 667 & 677 & 693 & 701 & 719 \\[0.5ex]
\textit{\textbf{Devanagari Numeral}} & 719 & 667 & 677 & 693 & 701 & 719 \\[0.5ex]
\textit{\textbf{Telugu Numeral}} & 719 & 667 & 677 & 693 & 701 & 719 \\[0.5ex]
\hline
\end{tabular}%
}
\end{table}

\begin{table}[thb]
\centering
\caption{Maximum Batch-size Possible for Numeral Datasets}
\label{tab:batch-size}
\resizebox{0.48\textwidth}{!}{%
\begin{tabular}{lcccccc}
\hline
\multirow{2}{*}{\textbf{Dataset}} & \multirow{2}{*}{\textbf{\begin{tabular}[c]{@{}c@{}}Capsule\\ Network\end{tabular}}} & \multicolumn{5}{c}{\textbf{$N_{features}$}} \\[0.5ex]
 &  & \textbf{2} & \textbf{4} & \textbf{6} & \textbf{8} & \textbf{10} \\\hline\hline 
\textit{\textbf{Bangla Numeral}} & 432 & 1520 & 960 & 694 & 534 & 432 \\[0.5ex]
\textit{\textbf{Devanagari Numeral}} & 432 & 1520 & 960 & 694 & 534 & 432 \\[0.5ex]
\textit{\textbf{Telugu Numeral}} & 432 & 1520 & 960 & 694 & 534 & 432 \\[0.5ex]
\hline
\end{tabular}%
}
\end{table}

\begin{table}[thb]
\centering
\caption{Accuracy (in percentage) of Proposed Approach on Numeral Datasets}
\label{tab:accuracy-numeral}
\resizebox{0.48\textwidth}{!}{%
\begin{tabular}{lcccccc}
\hline
\multirow{2}{*}{\textbf{Dataset}} & \multirow{2}{*}{\textbf{\begin{tabular}[c]{@{}c@{}}Capsule\\ Network\end{tabular}}} & \multicolumn{5}{c}{\textbf{$N_{features}$}} \\[0.5ex]
 &  & \textbf{2} & \textbf{4} & \textbf{6} & \textbf{8} & \textbf{10} \\\hline\hline 
\textit{\textbf{Bangla Numeral}} & \textbf{97.4} & 97.1 & 97.2 & 97.1 & 97.3 & 97.3 \\[0.5ex]
\textit{\textbf{Devanagari Numeral}} & \textbf{94.8} & 92.4 & \textbf{94.8} & 94.6 & \textbf{94.8} & 93.3 \\[0.5ex]
\textit{\textbf{Telugu Numeral}} & 96.2 & \textbf{96.9} & 96.4 & 95.9 & 96.2 & 95.3\\[0.5ex]
\hline
\end{tabular}%
}
\end{table}

\begin{table}[thb]
\centering
\caption{Overall Training Time (in seconds) per Epoch with Highest Possible Batch Size for Numeral Datasets}
\label{tab:net-time}
\resizebox{0.48\textwidth}{!}{%
\begin{tabular}{lcccccc}
\hline
\multirow{2}{*}{\textbf{Dataset}} & \multirow{2}{*}{\textbf{\begin{tabular}[c]{@{}c@{}}Capsule\\ Network\end{tabular}}} & \multicolumn{5}{c}{\textbf{$N_{features}$}} \\[0.5ex]
 &  & \textbf{2} & \textbf{4} & \textbf{6} & \textbf{8} & \textbf{10} \\\hline\hline 
\textit{\textbf{Bangla Numeral}} & 11.57 & 5.54 & 6.97 & 8.34 & 9.66 & 11.19 \\[0.5ex]
\textit{\textbf{Devanagari Numeral}} & 6.05 & 3.17 & 3.90 & 4.55 & 5.18 & 5.93 \\[0.5ex]
\textit{\textbf{Telugu Numeral}} & 6.18 & 3.16 & 3.94 & 4.57 & 5.18 & 5.94 \\[0.5ex]
\hline
\end{tabular}%
}
\end{table}

\begin{table*}[thb]
\centering
\caption{Training Time (in seconds) per Sample per Epoch for Higher-class Datasets}
\label{tab:higher-class-time-per-sample}
\resizebox{0.75\textwidth}{!}{%
\begin{tabular}{lcccccccc}
\hline
\multirow{2}{*}{\textbf{Datasets}} & \multirow{2}{*}{\textbf{\begin{tabular}[c]{@{}c@{}}Capsule \\ Network\end{tabular}}} & \multicolumn{7}{c}{\textbf{$N_{features}$}} \\
 &  & \textbf{2} & \textbf{4} & \textbf{6} & \textbf{8} & \textbf{10} & \textbf{15} & \textbf{20} \\[0.5ex] \hline \hline
\textit{\textbf{Bangla Basic}} & 0.0783 & 0.0660 & 0.0665 & 0.0671 & 0.0678 & 0.0681 & 0.0683 & 0.0698 \\[0.5ex]
\textit{\textbf{Bangla Compound}} & 0.0823 & 0.0662 & 0.0668 & 0.0672 & 0.0680 & 0.0684 & 0.0684 & 0.0699
\\[0.5ex]
\hline
\end{tabular}%
}
\end{table*}

\begin{table*}[thb]
\centering
\caption{Memory Consumption (in megabytes) per Sample for Higher-class Datasets}
\label{tab:higher-class-memory-consumption}
\resizebox{0.6\textwidth}{!}{%
\begin{tabular}{lcccccccc}
\hline
\multirow{2}{*}{\textbf{Datasets}} & \multirow{2}{*}{\textbf{\begin{tabular}[c]{@{}c@{}}Capsule \\ Network\end{tabular}}} & \multicolumn{7}{c}{\textbf{$N_{features}$}} \\
 &  & \textbf{2} & \textbf{4} & \textbf{6} & \textbf{8} & \textbf{10} & \textbf{15} & \textbf{20} \\[0.5ex] \hline \hline
\textit{\textbf{Bangla Basic}} & 1001 & 667 & 677 & 693 & 701 & 719 & 763 & 801 \\[0.5ex]
\textit{\textbf{Bangla Compound}} & 2057 & 667 & 677 & 693 & 701 & 719 & 763 & 801\\[0.5ex]
\hline
\end{tabular}%
}
\end{table*}

\begin{table*}[thb]
\centering
\caption{Maximum Batch-size Possible for Higher-class Datasets}
\label{tab:higher-class-batch-size}
\resizebox{0.6\textwidth}{!}{%
\begin{tabular}{lcccccccc}
\hline
\multirow{2}{*}{\textbf{Datasets}} & \multirow{2}{*}{\textbf{\begin{tabular}[c]{@{}c@{}}Capsule \\ Network\end{tabular}}} & \multicolumn{7}{c}{\textbf{$N_{features}$}} \\
 &  & \textbf{2} & \textbf{4} & \textbf{6} & \textbf{8} & \textbf{10} & \textbf{15} & \textbf{20} \\[0.5ex] \hline \hline
\textit{\textbf{Bangla Basic}} & 70 & 1520 & 960 & 694 & 534 & 432 & 290 & 172 \\[0.5ex]
\textit{\textbf{Bangla Compound}} & 16 & 1520 & 960 & 694 & 534 & 432 & 290 & 172\\[0.5ex]
\hline
\end{tabular}%
}
\end{table*}
\subsection{Datasets}

We have used five Indic handwritten datasets\footnote{https://code.google.com/archive/p/cmaterdb/downloads} for our experiments. Of them, Bangla numeral(CMATERdb 3.1.1), Devanagari numeral(CMATERdb 3.2.1) and Telugu numeral(CMATERdb 3.4.1) datasets have 10 classes each and are divided in the ratio 2:1 with two parts train and one part test. datasets named Bangla basic character(CMATERdb 3.1.2) and Bangla compound character(3.1.3.3) gives us 50 classes and 199 classes respectively (Table \ref{tab:Dataset}). These are the higher class problems that have considerable amount of time and memory to train on original proposed model of capsule network.

\subsection{Experimentation}

The proposed approach is benchmarked against the traditional capsule network on the five above mentioned datasets. The main goal of the experimentation is to demonstrate the efficiency of the proposed network for problems with large number of classes. The primary factor controlling the performance for the proposed approach is the number of features which is previously represented as $N_{features}$. The three numeral datasets namely Bangla, Devanagari and Telugu have 10 classes each. We have measured the performance of the proposed approach for $N_{features} \in [2,4,6,8,10]$. There was no point going higher than $10$ because as the speed of computation starts to drop below the speed traditional capsule networks as $N_{features}$ exceeds $N_{class}$.
For the higher class datasets such as the Bangla basic characters or Bangla compound characters, the proposed approach is tested for $N_{features} \in [2,4,6,8,10,15,20]$. The proposed network started to show saturation towards for $N_{features}$ higher than $20$. The model with best training performance was saved and performance of the model on the test set has been reported. Separate validation set was not used because dedicated regularization techniques are already implemented in the process. Along with accuracy, many other factors are also measured corresponding to the time and memory consumption of the network. All the experiments were conducted using a Nvidia GeForce GTX 1080ti(11GB).

\begin{table}[thb]
\centering
\caption{Accuracy (in percentage) of Proposed Approach on Higher-class Datasets}
\label{tab:higher-class-accuracy}
\resizebox{0.48\textwidth}{!}{%
\begin{tabular}{lcccccccc}
\hline
\multirow{2}{*}{\textbf{Datasets}} & \multirow{2}{*}{\textbf{\begin{tabular}[c]{@{}c@{}}Capsule \\ Network\end{tabular}}} & \multicolumn{7}{c}{\textbf{$N_{features}$}} \\
 &  & \textbf{2} & \textbf{4} & \textbf{6} & \textbf{8} & \textbf{10} & \textbf{15} & \textbf{20} \\[0.5ex] \hline \hline
\textit{\textbf{Bangla Basic}} & 90.6 & 91.07 & 91.73 & 92.7 & \textbf{93.23} & 92.1 & 92.97 & 92.1 \\[0.5ex]
\textit{\textbf{Bangla Compound}} & 79.34 & 79 & 85.53 & \textbf{87.44} & 87.11 & 87.04 & 86.54 & 87.24 \\[0.5ex]
\hline
\end{tabular}%
}
\end{table}

\begin{table}[thb]
\centering
\caption{Overall Training Time (in seconds) per Epoch with Highest Possible Batch Size for Higher-class Datasets}
\label{tab:higher-class-time}
\resizebox{0.48\textwidth}{!}{%
\begin{tabular}{lcccccccc}
\hline
\multirow{2}{*}{\textbf{Datasets}} & \multirow{2}{*}{\textbf{\begin{tabular}[c]{@{}c@{}}Capsule \\ Network\end{tabular}}} & \multicolumn{7}{c}{\textbf{$N_{features}$}} \\
 &  & \textbf{2} & \textbf{4} & \textbf{6} & \textbf{8} & \textbf{10} & \textbf{15} & \textbf{20} \\[0.5ex] \hline \hline
\textit{\textbf{Bangla Basic}} & 120.61 & 14.15 & 18.19 & 22.03 & 25.74 & 29.87 & 40.40 & 50.727 \\[0.5ex]
\textit{\textbf{Bangla Compound}} & 1206.99 & 36.92 & 47.57 & 59.65 & 70.11 & 81.91 & 110.75 & 149.58 \\[0.5ex]
\hline
\end{tabular}%
}
\end{table}

\subsection{Results and Discussion}

The first set of experiments show the performance of the network for numeral datasets. Though there is not much improvement in terms of accuracy (Table \ref{tab:accuracy-numeral}), the overall training time has been greatly reduced (Table \ref{tab:net-time}).

While the proposed network itself reduces the computational time for each sample (Table \ref{tab:time-per-sample}), the major speedup comes from the decrease in the memory consumption (Table \ref{tab:memory-consumption}). With a lower memory consumption it is possible to train the network with a higher batch size (Table \ref{tab:batch-size}).

The second set of experiments represents the scalability of the model. The proposed network is run on two datasets with higher number of classes namely, Bangla basic characters with 50 classes and Bangla compound character with 199 classes. The proposed model obtained a much better accuracy (Table \ref{tab:higher-class-accuracy}) for both cases with a much lesser training time (Table \ref{tab:higher-class-time}).

The training time for individual samples per epoch (Table \ref{tab:higher-class-time-per-sample}) shows a trend similar to the numeric datasets. Thus providing some boost in time consumption over basic capsule networks. However, the real benefit is in terms of the GPU memory consumption (Table \ref{tab:higher-class-memory-consumption}). Unlike capsule networks where the memory consumption increases with the increasing number of classes, the proposed method is immune to the problem. Because the dynamic routing for a fixed dimension of features is independent of the number of classes, the memory consumption is almost identical. The only increase in memory is due to the last fully connected layer from the feature capsules to the output layer, however the difference is negligible with respect to the overall consumption. For example, for a 10-dimensional feature capsules, extra memory requirement for the last fully connected layer for digit datasets would be 0.006 MiB, for Bangla basic dataset it is 0.031 MiB, and for Bangla compound character dataset it is 0.123 MiB. Hence it can be seen the difference is insignificant with respect to the total consumption. This allows the use of same batch size accross all datasets with different number of classes (Table \ref{tab:higher-class-batch-size}).

\begin{table}[thb]
\centering
\caption{Performance comparison with other approaches}
\label{tab:compare}
\resizebox{0.48\textwidth}{!}{%
\begin{tabular}{lcll}
\hline
\multirow{2}{*}{\textbf{Dataset}} & \multirow{2}{*}{\textbf{\begin{tabular}[c]{@{}c@{}}Proposed Method\\ ($N_{features}$)\end{tabular}}} & \multicolumn{2}{c}{\textbf{Other methods}} \\
 &  & \textbf{Method} & \textbf{Accuracy} \\ \hline
\textit{\textbf{Bangla}} & 97.3 (8) & LeNet~\cite{lecun1998gradient} & 94.6 \\
\textit{\textbf{Digits}} &  & CapsNet~\cite{sabour2017dynamic} & 97.4 \\
\textit{\textbf{}} &  & LeNet+CapsNet~\cite{mandal2018capsule} & 95.45 \\
\textit{\textbf{}} &  & Basu et al.~\cite{basu2012mlp} & 96.67 \\
\textit{\textbf{}} &  & Roy et al.~\cite{roy2012new} & 95.08 \\ \hline
\textit{\textbf{Devanagari}} & 94.8 (4) & LeNet~\cite{lecun1998gradient} & 92.1 \\
\textit{\textbf{Digits}} &  & CapsNet~\cite{sabour2017dynamic} & 94.8 \\
\textit{\textbf{}} &  & LeNet+CapsNet~\cite{mandal2018capsule} & 94 \\
\textit{\textbf{}} &  & Das et al.~\cite{das2010handwritten} & 90.44 \\ \hline
\textit{\textbf{Telugu}} & 96.9 (2) & LeNet~\cite{lecun1998gradient} & 95.8 \\
\textit{\textbf{Digits}} &  & CapsNet~\cite{sabour2017dynamic} & 96.2 \\
\textit{\textbf{}} &  & LeNet+CapsNet~\cite{mandal2018capsule} & 96.2 \\
\textit{\textbf{}} &  & Roy et al.~\cite{roy2014axiomatic} & 87.2 \\ \hline
\textit{\textbf{Bangla}} & 93.23 (8) & LeNet~\cite{lecun1998gradient} & 63 \\
\textit{\textbf{Basic}} &  & CapsNet~\cite{sabour2017dynamic} & 90.6 \\
\textit{\textbf{Characters}} &  & LeNet+CapsNet~\cite{mandal2018capsule} & 88.4 \\
\textit{\textbf{}} &  & Sarkhel et al.~\cite{sarkhel2015enhanced} & 86.53 \\
\textit{\textbf{}} &  & Bhattacharya et al~\cite{bhattacharya2006recognition}. & 92.15 \\ \hline
\textit{\textbf{Bangla}} & 87.44 (6) & LeNet~\cite{lecun1998gradient} & 75.4 \\
\textit{\textbf{Compound}} &  & CapsNet~\cite{sabour2017dynamic} & 79.3 \\
\textit{\textbf{Characters}} &  & LeNet+CapsNet~\cite{mandal2018capsule} & 79.9 \\
\textit{\textbf{}} &  & Sarkhel et al.~\cite{sarkhel2016multi} & 86.64 \\ \hline
\end{tabular}%
}
\end{table}
In several occassions we can see in table \ref{tab:compare} the proposed approach beating the classic capsule networks. The difference is more prominent for problems with higher number of classes like Bangla Basic and Bangla Compound Characters

\section{Conclusion}
One of the primary concerns of capsule network is its poor scalability for problems of larger classes. In our approach we have shown how capsule networks can be used to extract feature specific capsules rather than class specific capsules. Through this method a considerable boost has been demonstrated in terms of overall training time per epoch. Moreover the memory requirements of the proposed network is independent of the number of classes, thus allowing networks to be trained using a much higher number of batches and which provides a much more efficient implementation of the parallelization capabilities of a GPU. Agreement among capsule become much more complicated for higher class problems. Thus the normal capsule network tend to overfit the data. By generating intermediate features of lower dimensions, a more generalized learning environment is created. This results in an improvement in accuracy for higher class problems. The proposed network beats many other popular works on the current Indic datasests. In future, more analysis needs to be carried out regarding the effect of feature capsule on the concept of equivariance. Moreover, since reconstruction is carried out based on entire set of feature capsules, class specific reconstruction is not possible and hence provides another avenue to expand the work.

\section*{Acknowledgment}
This work is partially supported by the project order no. SB/S3/EECE/054/2016, dated 25/11/2016, sponsored by SERB (Government of India) and carried out at the Centre for Microprocessor Application for Training Education and Research, CSE Department, Jadavpur University. 

\ifCLASSOPTIONcaptionsoff
\newpage
\fi
\bibliographystyle{IEEETran}
\bibliography{ref}

\begin{thebibliography}{10}
\providecommand{\url}[1]{#1}
\csname url@samestyle\endcsname
\providecommand{\newblock}{\relax}
\providecommand{\bibinfo}[2]{#2}
\providecommand{\BIBentrySTDinterwordspacing}{\spaceskip=0pt\relax}
\providecommand{\BIBentryALTinterwordstretchfactor}{4}
\providecommand{\BIBentryALTinterwordspacing}{\spaceskip=\fontdimen2\font plus
\BIBentryALTinterwordstretchfactor\fontdimen3\font minus
  \fontdimen4\font\relax}
\providecommand{\BIBforeignlanguage}[2]{{%
\expandafter\ifx\csname l@#1\endcsname\relax
\typeout{** WARNING: IEEEtran.bst: No hyphenation pattern has been}%
\typeout{** loaded for the language `#1'. Using the pattern for}%
\typeout{** the default language instead.}%
\else
\language=\csname l@#1\endcsname
\fi
#2}}
\providecommand{\BIBdecl}{\relax}
\BIBdecl

\bibitem{lecun1998gradient}
Y.~LeCun, L.~Bottou, Y.~Bengio, and P.~Haffner, ``Gradient-based learning
  applied to document recognition,'' \emph{Proceedings of the IEEE}, vol.~86,
  no.~11, pp. 2278--2324, 1998.

\bibitem{krizhevsky2012imagenet}
A.~Krizhevsky, I.~Sutskever, and G.~E. Hinton, ``Imagenet classification with
  deep convolutional neural networks,'' in \emph{Advances in neural information
  processing systems}, 2012, pp. 1097--1105.

\bibitem{roy2017handwritten}
S.~Roy, N.~Das, M.~Kundu, and M.~Nasipuri, ``Handwritten isolated bangla
  compound character recognition: A new benchmark using a novel deep learning
  approach,'' \emph{Pattern Recognition Letters}, vol.~90, pp. 15--21, 2017.

\bibitem{szegedy2015going}
C.~Szegedy, W.~Liu, Y.~Jia, P.~Sermanet, S.~Reed, D.~Anguelov, D.~Erhan,
  V.~Vanhoucke, and A.~Rabinovich, ``Going deeper with convolutions,'' in
  \emph{Proceedings of the IEEE conference on computer vision and pattern
  recognition}, 2015, pp. 1--9.

\bibitem{he2016deep}
K.~He, X.~Zhang, S.~Ren, and J.~Sun, ``Deep residual learning for image
  recognition,'' in \emph{Proceedings of the IEEE conference on computer vision
  and pattern recognition}, 2016, pp. 770--778.

\bibitem{huang2017densely}
G.~Huang, Z.~Liu, L.~Van Der~Maaten, and K.~Q. Weinberger, ``Densely connected
  convolutional networks.'' in \emph{CVPR}, vol.~1, no.~2, 2017, p.~3.

\bibitem{sabour2017dynamic}
S.~Sabour, N.~Frosst, and G.~E. Hinton, ``Dynamic routing between capsules,''
  in \emph{Advances in Neural Information Processing Systems}, 2017, pp.
  3856--3866.

\bibitem{mandal2018capsule}
B.~Mandal, S.~Dubey, R.~Sarkhel, and N.~Das, ``Handwritten indic character
  recognition using capsule networks,'' in \emph{Proceedings of the 1st IEEE
  Conference on Applied Signal Processing}, 2018.

\bibitem{hinton2018matrix}
G.~E. Hinton, S.~Sabour, and N.~Frosst, ``Matrix capsules with em routing,''
  2018.

\bibitem{xi2017capsule}
E.~Xi, S.~Bing, and Y.~Jin, ``Capsule network performance on complex data,''
  \emph{arXiv preprint arXiv:1712.03480}, 2017.

\bibitem{obaidullah2018phdindic_11}
S.~M. Obaidullah, C.~Halder, K.~Santosh, N.~Das, and K.~Roy, ``Phdindic\_11:
  page-level handwritten document image dataset of 11 official indic scripts
  for script identification,'' \emph{Multimedia Tools and Applications},
  vol.~77, no.~2, pp. 1643--1678, 2018.

\bibitem{basu2012mlp}
S.~Basu, N.~Das, R.~Sarkar, M.~Kundu, M.~Nasipuri, and D.~K. Basu, ``An mlp
  based approach for recognition of handwrittenbangla'numerals,'' \emph{arXiv
  preprint arXiv:1203.0876}, 2012.

\bibitem{roy2012new}
A.~Roy, N.~Mazumder, N.~Das, R.~Sarkar, S.~Basu, and M.~Nasipuri, ``A new quad
  tree based feature set for recognition of handwritten bangla numerals,'' in
  \emph{Engineering Education: Innovative Practices and Future Trends (AICERA),
  2012 IEEE International Conference on}.\hskip 1em plus 0.5em minus
  0.4em\relax IEEE, 2012, pp. 1--6.

\bibitem{das2010handwritten}
N.~Das, B.~Das, R.~Sarkar, S.~Basu, M.~Kundu, and M.~Nasipuri, ``Handwritten
  bangla basic and compound character recognition using mlp and svm
  classifier,'' \emph{arXiv preprint arXiv:1002.4040}, 2010.

\bibitem{roy2014axiomatic}
A.~Roy, N.~Das, R.~Sarkar, S.~Basu, M.~Kundu, and M.~Nasipuri, ``An axiomatic
  fuzzy set theory based feature selection methodology for handwritten numeral
  recognition,'' in \emph{ICT and Critical Infrastructure: Proceedings of the
  48th Annual Convention of Computer Society of India-Vol I}.\hskip 1em plus
  0.5em minus 0.4em\relax Springer, 2014, pp. 133--140.

\bibitem{sarkhel2015enhanced}
R.~Sarkhel, A.~K. Saha, and N.~Das, ``An enhanced harmony search method for
  bangla handwritten character recognition using region sampling,'' in
  \emph{Recent Trends in Information Systems (ReTIS), 2015 IEEE 2nd
  International Conference on}.\hskip 1em plus 0.5em minus 0.4em\relax IEEE,
  2015, pp. 325--330.

\bibitem{bhattacharya2006recognition}
U.~Bhattacharya, M.~Shridhar, and S.~K. Parui, ``On recognition of handwritten
  bangla characters,'' in \emph{Computer Vision, Graphics and Image
  Processing}.\hskip 1em plus 0.5em minus 0.4em\relax Springer, 2006, pp.
  817--828.

\bibitem{sarkhel2016multi}
R.~Sarkhel, N.~Das, A.~K. Saha, and M.~Nasipuri, ``A multi-objective approach
  towards cost effective isolated handwritten bangla character and digit
  recognition,'' \emph{Pattern Recognition}, vol.~58, pp. 172--189, 2016.

\end{thebibliography}

\end{document}